\pdfoutput=1

\documentclass[11pt]{article}
\usepackage{booktabs}
\usepackage{multirow}
\usepackage{subfigure}
\usepackage{subcaption}
\usepackage{acl}
\usepackage{array}
\usepackage{longtable}
\usepackage{geometry}
\usepackage{times}
\usepackage{latexsym}
\usepackage{graphicx}
\usepackage{adjustbox}
\usepackage{float}
\usepackage{nicematrix}
\usepackage{siunitx}
\usepackage{bm}
\usepackage{soul}
\usepackage{booktabs}
\usepackage{lscape}
\usepackage{longtable}
\usepackage{CJKutf8}

\usepackage[T1]{fontenc}

\usepackage[utf8]{inputenc}

\usepackage{microtype}

\usepackage{xcolor}
\usepackage{colortbl}
\usepackage{xspace}
\usepackage{ifthen}

\definecolor{Ind}{HTML}{F17D7D}
\definecolor{US}{HTML}{3C6C76}
\definecolor{Trans}{HTML}{9A70ED}
\definecolor{Gen}{HTML}{458657}
\definecolor{Red}{HTML}{AC2037}
\definecolor{Gray}{gray}{0.8}
\definecolor{filler}{HTML}{9F080E}
\definecolor{rephrase}{HTML}{F7AB1E}
\definecolor{mask}{HTML}{9F080E}
\definecolor{prune}{HTML}{9F080E}
\newcolumntype{a}{>{\columncolor{Gray}}c}

\newboolean{usecolor}
\setboolean{usecolor}{true}

\newcommand{\IndEng}{%
\ifthenelse{\boolean{usecolor}}%
{\texttt{\textbf{\textcolor{Ind}{IndEng}}}\xspace}%
{\texttt{\textbf{IndEng}}\xspace}%
}
\newcommand{\USEng}{%
\ifthenelse{\boolean{usecolor}}%
{\texttt{\textbf{\textcolor{US}{USEng}}}}%
{\texttt{\textbf{USEng}}}%
\xspace%
}
\newcommand{\AITrans}{%
\ifthenelse{\boolean{usecolor}}%
{\texttt{\textbf{\textcolor{Trans}{AITrans}}}\xspace}%
{\texttt{\textbf{AITrans}}\xspace}%
}
\newcommand{\AIGen}{%
\ifthenelse{\boolean{usecolor}}%
{\texttt{\textbf{\textcolor{Gen}{AIGen}}}\xspace}%
{\texttt{\textbf{AIGen}}\xspace}%
}

\usepackage{inconsolata}
\usepackage{cleveref}

\crefformat{section}{\S#2#1#3}
\crefformat{subsection}{\S#2#1#3}
\crefformat{subsubsection}{\S#2#1#3}

\usepackage{amssymb}
\usepackage{pifont}

\usepackage{amsmath}

\newcounter{eqn}

\makeatletter
\newcommand{\putindeepbox}[2][0.7\baselineskip]{{%
    \setbox0=\hbox{#2}%
    \setbox0=\vbox{\noindent\hsize=\wd0\unhbox0}
    \@tempdima=\dp0
    \advance\@tempdima by \ht0
    \advance\@tempdima by -#1\relax
    \dp0=\@tempdima
    \ht0=#1\relax
    \box0
}}
\makeatother

\title{``Is Hate Lost in Translation?'': Evaluation of Multilingual LGBTQIA+ Hate Speech Detection}

\author{Fai Leui Chan, Duke Nguyen, Aditya Joshi\\
University of New South Wales, Sydney, Australia\\
\quad\texttt{aditya.joshi@unsw.edu.au} \\}

\begin{document}
\maketitle
\begin{abstract}

This paper explores the challenges of detecting LGBTQIA+ hate speech of large language models across multiple languages, including English, Italian, Chinese and (code-switched) English-Tamil, examining the impact of machine translation and whether the nuances of hate speech are preserved across translation. We examine the hate speech detection ability of zero-shot and fine-tuned GPT. Our findings indicate that: (1) English has the highest performance and the code-switching scenario of English-Tamil being the lowest, (2) fine-tuning improves performance consistently across languages whilst translation yields mixed results. Through simple experimentation with original text and machine-translated text for hate speech detection along with a qualitative error analysis, this paper sheds light on the socio-cultural nuances and complexities of languages that may not be captured by automatic translation.

\begin{table*}[t!]
\centering
\scalebox{0.85}{
\begin{tabular}{lllll}
\toprule
\textbf{Language} & \textbf{Source} & \textbf{Total Samples} & \textbf{Non-Homotransphobic} & \textbf{Homotransphobic} \\ \midrule
English       & \citep{mcgiff2024bridging}   & 1,277 & 656 (51.4\%)   & 621 (48.6\%)   \\
Italian       & \citep{HODI}               & 5,000 & 2,992 (59.8\%) & 2,008 (40.2\%) \\
Chinese       & \citep{toxicn}             & 2011  & 1247 (62.0\%)  & 764 (38.0\%)   \\
English-Tamil & \citep{chakravarthi2021dataset} & 6033  & 5384 (89.2\%)  & 649 (10.8\%)   \\
\bottomrule
\end{tabular}%
}
\caption{Comparison of datasets. \% in the Non-Homotransphobic and Homotransphobic columns refer to the proportion of each class relative to the total samples in each dataset, with each row summing to 100\%.}
\label{table:comparison-datasets}
\end{table*}

\end{abstract}
\textbf{Warning: The paper contains examples of multilingual hate speech towards LGBTQIA+ community because of the nature of the work.}
\section{Introduction}
LGBTQIA+ individuals are particularly vulnerable to hate speech due to their sexual orientation and gender identity. They are frequently subject to harassment, discrimination, violence due to their identity \citep{lt-edi-2024}. Therefore, many social media platforms have implemented hate speech detection as part of content sanitation on their platforms to create safer online environments. As social media platforms become increasingly diverse with people coming from different linguistic backgrounds, we investigate if hate speech detection is sustained across different languages, translations, and code-switching environments. In other words, is hate speech detection ``lost in translation''\footnote{As part of a discussion on his poem ``Stopping by Woods on a Snowy Evening'', Robert Frost famously remarked ``You've often heard me say – perhaps too often – that poetry is what is lost in translation. It is also what is lost in interpretation.'' \citep[p. 18]{untermeyer1964robert}}?


\begin{figure}[t!]
\centering
    \includegraphics[width=0.9\linewidth]{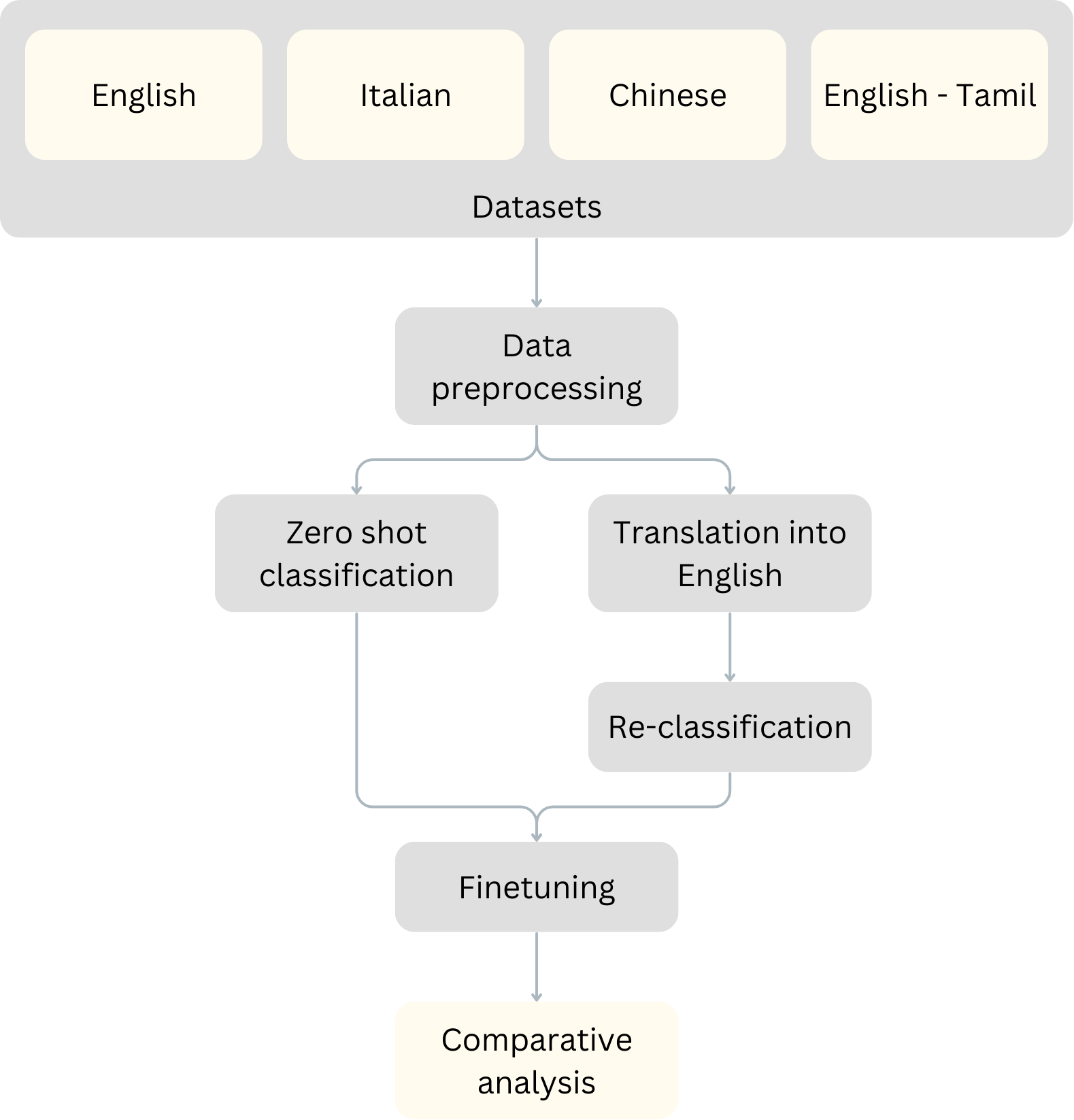}
    \caption{Evaluation Methodology of Machine Translation-based Hate Speech Detection.}
    \label{fig:evaluation-process}
\end{figure}

The approach of using machine translation to translate the test data into English and running inference using an English-only model has long been studied~\cite{pikuliak2021cross}. This method may be better for complex tasks that require common sense or real-world knowledge, as it benefits from the use of a stronger English-only model \citep{artetxe-etal-2023-revisiting}, which may be useful for the complex task of hate speech detection.


Therefore, we ask the question: ``How does hate speech detection perform for original text and translated text?''
We do so for the case of hate speech towards LGBTQIA+ people. While it is intuitive that machine translation will not preserve all semantics, our experiments with zero-shot and fine-tuned GPT show that it particularly holds true for hate speech detection. Our error analysis sheds light on the nature of errors to highlight `what' is lost in translation.
\section{Methodology}
Our methodology is as shown in Figure \ref{fig:evaluation-process}. We utilise labeled datasets in English, Italian, Chinese, and English-Tamil (code-mixed), each focusing on LGBTQIA+-specific hate speech. Our preprocessing involves removing excess spaces and invalid characters.

\begin{table*}[h!]
\centering
\scalebox{0.82}{
\begin{tabular}{@{}p{2.5cm}p{4cm}p{1.5cm}p{1.5cm}p{1.5cm}p{1.5cm}p{1.5cm}p{1.5cm}p{1.5cm}p{1.5cm}@{}}
\toprule
\textbf{Language} & \textbf{Condition} & \textbf{F1} & \textbf{P} & \textbf{R} & \textbf{K} & \textbf{$\Delta$ F} & \textbf{$\Delta$ K} \\ \midrule
English       & Original                  & 0.7952 & 0.7082 & 0.9066 & 0.5488 & -       & -       \\
              & Fine-tuned                & 0.8689 & 0.8833 & 0.8548 & 0.7486 & +0.0737 & +0.1998 \\
Italian       & Original                  & 0.5990 & 0.4514 & 0.8899 & 0.1414 & -       & -       \\
              & Translated                & 0.5355 & 0.4424 & 0.6783 & 0.0960 & -0.0635 & -0.0454 \\
              & Fine-tuned   (Original)   & 0.8375 & 0.8417 & 0.8333 & 0.7292 & +0.2385 & +0.5878 \\
              & Fine-tuned   (Translated) & 0.7417 & 0.7371 & 0.7463 & 0.5662 & +0.2062 & +0.4702 \\
Chinese       & Original                  & 0.7464 & 0.7493 & 0.7435 & 0.5878 & -       & -       \\
              & Translated                & 0.6839 & 0.7099 & 0.6597 & 0.2463 & -0.0625 & -0.3415 \\
              & Fine-tuned   (Original)   & 0.8146 & 0.8255 & 0.8039 & 0.7030 & +0.0682 & +0.1152 \\
              & Fine-tuned   (Translated) & 0.7661 & 0.7958 & 0.7386 & 0.6308 & +0.0822 & +0.3845 \\ 
English- Tamil & Original                  & 0.3619 & 0.2843 & 0.4977 & 0.1998 & -       & -       \\
              & Translated                & 0.3202 & 0.3511 & 0.2943 & 0.2463 & -0.0417 & +0.0465 \\
              & Fine-tuned   (Original)   & 0.5391 & 0.6200 & 0.4769 & 0.2452 & +0.1772 & +0.0454 \\
              & Fine-tuned   (Translated) & 0.4037 & 0.5000 & 0.3385 & 0.3469 & +0.0835 & +0.1006 \\
              \bottomrule
\end{tabular}
}
\caption{Performance Metrics (F1: F1-score, P: Precision, R: Recall: K: Cohen's Kappa) and Changes Across Languages and Conditions. $\Delta$ columns represent the changes in F1-score and Cohen’s Kappa between different conditions: Fine-tuned (Original → Fine-tuned), Translated (Original → Translated), and Fine-tuned (Translated → Fine-tuned).}
\label{tab:performance}
\end{table*}

We first perform zero-shot classification using a LLM (large language model) to classify whether the sentence was homotransphobic\footnote{`Homotransphobic' is used as an umbrella term to indicate hate speech towards the LGBTQIA+ community}, with 1 referring to homotransphobic content and 0 referring to non-homotransphobic content. Then, all non-English sentences are translated into English via the LLM using the following prompt: `Translate this sentence into English: ‘{text}’'. After translation, we subject the English text to a second round of classification with the system prompt being ``You are an AI assistant that classifies text as either homotransphobic (1) or not homotransphobic (0). Respond with only 0 or 1.”, and the user prompt being “Classify the following text: ‘{text}". We then finetune and evaluate the LLM on selected datasets. Finally, we assess the impact of translation on the final effectiveness of the model and measure the performance improvement, if any, achieved through fine-tuning.

\section{Experiment Setup}
The LLM which we use for our experiments is the gpt-3.5-turbo model\footnote{\url{https://platform.openai.com/docs/models}}, a chat-bot based on the GPT-3.5 language model developed by OpenAI. This model is optimised for prompt-based usage but performs equally well for traditional NLP tasks \citep{das-etal-2024-evaluating}. 

The datasets which we employ are shown in Table \ref{table:comparison-datasets} with a train-validation-test split of 60:20:20. It is noted that the datasets display varying degrees of imbalance which could affect model performance across languages. While the English and Italian datasets are fairly balanced, and the Chinese dataset shows a moderate imbalance, the English-Tamil dataset exhibits severe imbalance, with only 10.8\% of samples being homotransphobic, broadly referred to as hate speech towards the LGBTQIA+ community. 

The downstream task is hate speech detection, and is evaluated using the following metrics: F1 score, precision, recall, and Cohen-Kappa agreement. In particular, Cohen's Kappa is used to measure the agreement between the predicted labels and the true label. It is chosen as it is a good measure of intra-rater reliability, while correcting for times when the raters may agree by chance \citep{cohen}.

\section{Results}

\begin{table*}[tp]
\begin{CJK}{UTF8}{gbsn}
\centering
\scalebox{0.59}{
\begin{tabular}{m{1.5cm} m{4cm} m{4cm} m{4cm} m{0.2cm} m{0.2cm} m{9cm}}
\toprule
\textbf{Language} & \textbf{Original Text}                                                           & \textbf{GPT Translation}                                                              & \textbf{Human Translation}                                                 & \textbf{T} & \textbf{P} & \textbf{Analysis}                                                                                                                                                      \\ \midrule
Chinese           & 你怎么骂基佬我也不生气我也不是男的我还有男朋友                                                          & I'm not angry even if you scold me, I'm not a man, I have a boyfriend.                & I won't be offended if you call me gay. I'm not a man. I have a boyfriend. & 0              & 1                   & Mistranslation of ``基佬'' (brash slang for gay people).                                                                                                                 \\ \midrule
Chinese           & 男同没皮没脸要借别人的好脸                                                                    & Men who have no shame want to borrow other people's good reputation.                  & Gay men who have no shame want to borrow other people's good reputation.   & 1              & 0                   & ``男同'' (abbreviation for gay men) was missed in translation.                                                                                                           \\ \midrule
Chinese           & 妇男，快12点了，你上飞机没有？                                                                 & Woman, it's almost 12 o'clock, have you boarded the plane?                            & House-husband, it's almost 12 o'clock, have you boarded the plane?         & 1              & 0                   & Mistranslation of ``妇男'' (house-husband) as woman.                                                                                                                     \\ \midrule
English-Tamil     & Yogesh Waran Hey nee 9 taane                                                     & Yogesh, you are a 9 out of 10                                                         & Yogesh Waran, aren't you trans?                                            & 1              & 0                   & Mistranslation of ``9'' (double meaning word for the transgender community).                                                                                           \\ \midrule
English-Tamil     & Serupa alla ADI naramunda                                                        & "Similar to ADI, naramunda"                                                           & Will beat you with slipper, fool                                           & 1              & 0                   & Translation is completely wrong. ``Serupu'' means ``slipper/footwear'' and ``naramunda'' is pejorative term meaning ``fool''.                                          \\ \midrule
English-Tamil     & Serpala adikanum......enga ponanulm ithuka tholla thanga mudila ...              & I need to talk to Serpala...I can't handle this on my own...                          & I will hit with slipper, I cant bear this trouble.                         & 1              & 0                   & Mistranslation of Serpala, which means ``with slipper/footwear'' with an informal and aggressive connotation, as a name.                                               \\ \midrule
Italian           & @user\_ab   @user\_abcde @user\_abcdef @user\_a Sono tutti innamorati del busone & @user\_ab @user\_abcde @user\_abcdef @user\_a   They are all in love with the big bus & They're all in love with anal sex                                          & 1              & 0                   & Mistranslation of ``Busone'' (Northern Italian slang for anal sex, used for mocking homosexuals) as ``big bus'' (-one affix is used to indicate something that is big) \\ \midrule
Italian           & MADONNA SUCCHIACAZZI E PUTTANA \#bestemmie   \#world                             & MADONNA SUCKS DICK AND WHORE \#blasphemy \#world                                      & GOD FUCKING DAMMIT                                                         & 0              & 1                   & Mistranslation of ``Madonna Puttana'' which is a common rage expression in Italian involving saints but is not homotransphobic                                         \\ \midrule
Italian           & Non ho letto nulla   ma nel dubbio Sala ricchione                                & I haven't read anything but just in case   Sala is a rich man                         & I haven't read anything but just in case   Sala is a faggot                & 1              & 0                   & Mistranslation of ``ricchione'' (Southern Italian derogatory slang for homosexual people) as ``rich'' (which is ``ricco'') \\ \bottomrule

\end{tabular}
}
\caption{Qualitative error analysis of misclassified examples. Each sample is given the `Original Text', the `GPT Translation', and the `Human Translation'. `T' stands for `Truth' and `P' stands for `Prediction'. `Truth' and 'Prediction' values are either 0 (non-homotransphobic) or 1 (homotransphobic). `Analysis' are comments on the translation error.}
\label{tab:error_analysis}
\end{CJK}
\end{table*}

\subsection{Quantitative Evaluation}

Table \ref{tab:performance} compares the performance of gpt3.5-turbo on original text versus translated text across different languages. English yields the highest F1-score (0.7952), followed by Chinese (0.7464), Italian (0.5990), and English-Tamil (0.3619). The strong performance in Chinese suggests good generalisation to non-Latin scripts after translation, while the low score for English-Tamil highlights challenges with code-mixed content \citep{dogruoz-etal-2021-survey}.

We also evaluate whether applying the subsequent transformation process degrades or improves the performance. Translating non-English content to English produces mixed results. English-Tamil sees a slight improvement in Cohen's Kappa (+0.0465) despite a decrease in F1-score (-0.0417), which suggests translating and classifying may improve model performance in code-mixed languages \citep{gautam-etal-2021-translate}. Italian shows marginal decreases in both metrics. Chinese experiences the most significant performance drop (F1: -0.0625, Kappa: -0.3415), suggesting substantial loss of context during translation. These findings indicate that in general, translation decreases the effectiveness of hate speech detection. However, the degree of reduction is language-dependent.

Fine-tuning consistently improves performance across all languages, with the most substantial gains in Italian ($\Delta$F1: +0.2385, $\Delta$Kappa: +0.5878) and English-Tamil ($\Delta$F1: +0.1772). Even English and Chinese, which have strong baseline performances, see notable improvements. Fine-tuning on translated text also shows benefits, though generally not as substantial as fine-tuning on original text, with Chinese being an exception.

\subsection{Qualitative Analysis}
\begin{CJK}{UTF8}{gbsn}
We now show qualitative analysis of how hate is `lost in translation' as shown in the previous section. This is visible in the case of slang and culturally specific references. We request the assistance of native speakers of Italian, Tamil, and Chinese to identify prominent translation errors for the misclassified case as shown in Table \ref{tab:error_analysis}.

Table \ref{tab:error_analysis} indicates that most of LGBTQ terminologies, derogatory language involving LGBTQ people, and sometimes even non-LGBTQ slang words (in the case of English-Tamil) are mistranslated across the three languages (Italian, English-Tamil, and Chinese). This suggests that while the translation models may handle standard language adequately, they struggle with specialised or sub-cultural terms, which are often crucial in detecting hate speech. Despite these challenges, the model shows some strengths, such as correctly identifying some LGBTQIA+-related slang like ``BL"， ``CP"， and ``腐女" in translations.

\end{CJK}

The qualitative analysis reveals significant challenges in translating and detecting LGBTQIA+ hate speech across languages, particularly with slang, implicit hate, and cultural-specific expressions. While the model shows promise in some areas, there is a clear need for more nuanced, language-specific approaches to improve accuracy in multilingual hate speech detection. 
\section{Related Work}
Despite broad interest in hate speech detection, research specifically addressing LGBTQIA+ communities remain limited. Challenges to create a generalised hate speech model for various targets have been reported in particular\citep{HODI}. Shared tasks have been particularly important for hate speech detection towards LGBTQIA+ community. The LT-EDI@EACL series (2022-2024) focuses on the identification of homophobia, transphobia, and nonanti-LGBTQIA+ content in Tamil, English, and code-mixed English-Tamil \citep{lt-edi-2022, lt-edi-2023, lt-edi-2024}. The shared task has expanded to include various languages to look at homotransphobia in a multilingual context. There have also been other shared tasks on the topic, focusing on various languages. Examples include HOMO-MEX2023@IberLEF which focuses on hate speech detection towards the Mexican Spanish-Speaking LGBTQIA+ population \citep{homo-mex, shahiki2023lidoma}. In a similar vein, HODI is a shared task for the automatic detection of homotransphobia in Italian presented at EVALITA 2023 \citep{HODI}. Beyond shared tasks, some research has employed Transformer-based models like BERT and XLMRoBERTa to identify transphobic and homophobic insults in social media comments \citep{manikandan2022system}. Benchmarks such as WinoQueer~\cite{felkner2023winoqueer} provide pairs of sentences to measure anti-LGBTQIA+ bias in language models. To the best of our knowledge, this is the first hate speech detection comparison centered around machine translation. The datasets we use are reported in past work.
\section{Conclusion}
This study provides valuable insights into the effectiveness of LLM in hate speech detection in diverse linguistic settings involving LGBTQIA+ communities. We compare the ability of zero-shot and fine-tuned GPT for hate speech detection of multilingual text in the original language and translated versions to English. Our insights were: (1) hate speech detection via LLM is in general effective (including in non-Latin script settings), however LLMs perform significantly worse when dealing with code-mixed languages; (2) hate speech detection via LLM can be improved simply via fine-tuning, although the degree of improvement is language-dependent; (3) translation is ineffective in transferring nuanced ideas and show visible degradation on hate speech detection performance. 

To the best of our knowledge, this is the first work in hate speech detection with machine translation as our anchor. While the technique itself is simplistic, our research demonstrates the complexity of hate speech detection, especially for LGBTQIA+ communities in multilingual contexts and the need for continued research in this area. By advancing our understanding of multilingual hate speech detection, we can work towards creating safer, more inclusive online spaces for LGBTQIA+ individuals across different linguistic communities.


\section*{Limitations}
We now discuss limitation of this work. First of all, large language models have shown to exhibit bias towards LGBTQIA+ communities \citep{sosto2024queerbenchquantifyingdiscriminationlanguage, felkner2023winoqueer}, and there may exist potential biases in the training data and model itself. Secondly, the cascaded approach of using gpt3.5-turbo for both translation and classification makes the process vulnerable to errors from both stages and may introduce biases or errors that are difficult to isolate \citep{unanue-etal-2023-t3l}. In addition, the use of GPT is prompt-dependant. The quality of the prompt can significantly impact the quality and accuracy of the model’s outputs~\citep{li2024}. Our works have not analyzed the effects of insignificant prompt variation on the model's performance on selected tasks. Furthermore, there is a lack of context beyond single sentences in our analysis. Providing more contextual information could lead to a more robust understanding of the cultural context and lead to better results. Moreover, we have not analyzed if there was any correlation between the translation quality and the performance on the downstream tasks. Lastly, it would be highly beneficial to compare gpt3.5-turbo with other large language models and specialised hate speech detection systems to benchmark its effectiveness. 
\section*{Acknowledgment}
This paper is the outcome of a Taste-of-Research scholarship awarded to Fai Leui Chan. The  scholarship was funded by Google's exploreCSR grant.

We would like to acknowledge Marsha Mariya Kappan and Steffano Mezza from the School of Computer Science and Engineering at the University of New South Wales for providing qualitative error analysis of English-Tamil and Italian classification examples.
\bibliography{refs}

\begin{thebibliography}{21}
\providecommand{\natexlab}[1]{#1}

\bibitem[{Artetxe et~al.(2023)Artetxe, Goswami, Bhosale, Fan, and Zettlemoyer}]{artetxe-etal-2023-revisiting}
Mikel Artetxe, Vedanuj Goswami, Shruti Bhosale, Angela Fan, and Luke Zettlemoyer. 2023.
\newblock \href {https://doi.org/10.18653/v1/2023.emnlp-main.399} {Revisiting machine translation for cross-lingual classification}.
\newblock In \emph{Proceedings of the 2023 Conference on Empirical Methods in Natural Language Processing}, pages 6489--6499, Singapore. Association for Computational Linguistics.

\bibitem[{Bel{-}Enguix et~al.(2023)Bel{-}Enguix, G{\'{o}}mez{-}Adorno, Sierra, V{\'{a}}squez, Andersen, and Ojeda{-}Trueba}]{homo-mex}
Gemma Bel{-}Enguix, Helena G{\'{o}}mez{-}Adorno, Gerardo Sierra, Juan V{\'{a}}squez, Scott~Thomas Andersen, and Sergio Ojeda{-}Trueba. 2023.
\newblock \href {http://journal.sepln.org/sepln/ojs/ojs/index.php/pln/article/view/6566} {Overview of {HOMO-MEX} at iberlef 2023: Hate speech detection in online messages directed towards the mexican spanish speaking {LGBTQ+} population}.
\newblock \emph{Proces. del Leng. Natural}, 71:361--370.

\bibitem[{Chakravarthi et~al.(2024)Chakravarthi, Kumaresan, Priyadharshini, Buitelaar, Hegde, Shashirekha, Rajiakodi, Garc{\'\i}a, Jim{\'e}nez-Zafra, Garc{\'\i}a-D{\'\i}az, Valencia-Garc{\'\i}a, Ponnusamy, Shetty, and Garc{\'\i}a-Baena}]{lt-edi-2024}
Bharathi~Raja Chakravarthi, Prasanna Kumaresan, Ruba Priyadharshini, Paul Buitelaar, Asha Hegde, Hosahalli Shashirekha, Saranya Rajiakodi, Miguel~{\'A}ngel Garc{\'\i}a, Salud~Mar{\'\i}a Jim{\'e}nez-Zafra, Jos{\'e} Garc{\'\i}a-D{\'\i}az, Rafael Valencia-Garc{\'\i}a, Kishore Ponnusamy, Poorvi Shetty, and Daniel Garc{\'\i}a-Baena. 2024.
\newblock \href {https://aclanthology.org/2024.ltedi-1.11} {Overview of third shared task on homophobia and transphobia detection in social media comments}.
\newblock In \emph{Proceedings of the Fourth Workshop on Language Technology for Equality, Diversity, Inclusion}, pages 124--132, St. Julian's, Malta. Association for Computational Linguistics.

\bibitem[{Chakravarthi et~al.(2023)Chakravarthi, Ponnusamy, S, Buitelaar, Garc{\'\i}a-Cumbreras, Jimenez-Zafra, Garcia-Diaz, Valencia-Garcia, and Jindal}]{lt-edi-2023}
Bharathi~Raja Chakravarthi, Rahul Ponnusamy, Malliga S, Paul Buitelaar, Miguel~{\'A}ngel Garc{\'\i}a-Cumbreras, Salud~Mar{\'\i}a Jimenez-Zafra, Jose~Antonio Garcia-Diaz, Rafael Valencia-Garcia, and Nitesh Jindal. 2023.
\newblock \href {https://aclanthology.org/2023.ltedi-1.6} {Overview of second shared task on homophobia and transphobia detection in social media comments}.
\newblock In \emph{Proceedings of the Third Workshop on Language Technology for Equality, Diversity and Inclusion}, pages 38--46, Varna, Bulgaria. INCOMA Ltd., Shoumen, Bulgaria.

\bibitem[{Chakravarthi et~al.(2022)Chakravarthi, Priyadharshini, Durairaj, McCrae, Buitelaar, Kumaresan, and Ponnusamy}]{lt-edi-2022}
Bharathi~Raja Chakravarthi, Ruba Priyadharshini, Thenmozhi Durairaj, John McCrae, Paul Buitelaar, Prasanna Kumaresan, and Rahul Ponnusamy. 2022.
\newblock \href {https://doi.org/10.18653/v1/2022.ltedi-1.57} {Overview of the shared task on homophobia and transphobia detection in social media comments}.
\newblock In \emph{Proceedings of the Second Workshop on Language Technology for Equality, Diversity and Inclusion}, pages 369--377, Dublin, Ireland. Association for Computational Linguistics.

\bibitem[{Chakravarthi et~al.(2021)Chakravarthi, Priyadharshini, Ponnusamy, Kumaresan, Sampath, Thenmozhi, Thangasamy, Nallathambi, and McCrae}]{chakravarthi2021dataset}
Bharathi~Raja Chakravarthi, Ruba Priyadharshini, Rahul Ponnusamy, Prasanna~Kumar Kumaresan, Kayalvizhi Sampath, Durairaj Thenmozhi, Sathiyaraj Thangasamy, Rajendran Nallathambi, and John~Phillip McCrae. 2021.
\newblock \href {https://arxiv.org/abs/2109.00227} {Dataset for identification of homophobia and transophobia in multilingual youtube comments}.
\newblock \emph{Preprint}, arXiv:2109.00227.

\bibitem[{Cohen(1960)}]{cohen}
Jacob Cohen. 1960.
\newblock \href {https://doi.org/10.1177/001316446002000104} {A coefficient of agreement for nominal scales}.
\newblock \emph{Educational and Psychological Measurement}, 20(1):37--46.

\bibitem[{Das et~al.(2024)Das, Pandey, and Mukherjee}]{das-etal-2024-evaluating}
Mithun Das, Saurabh~Kumar Pandey, and Animesh Mukherjee. 2024.
\newblock \href {https://aclanthology.org/2024.lrec-main.564} {Evaluating {C}hat{GPT} against functionality tests for hate speech detection}.
\newblock In \emph{Proceedings of the 2024 Joint International Conference on Computational Linguistics, Language Resources and Evaluation (LREC-COLING 2024)}, pages 6370--6380, Torino, Italia. ELRA and ICCL.

\bibitem[{Do{\u{g}}ru{\"o}z et~al.(2021)Do{\u{g}}ru{\"o}z, Sitaram, Bullock, and Toribio}]{dogruoz-etal-2021-survey}
A.~Seza Do{\u{g}}ru{\"o}z, Sunayana Sitaram, Barbara~E. Bullock, and Almeida~Jacqueline Toribio. 2021.
\newblock \href {https://doi.org/10.18653/v1/2021.acl-long.131} {A survey of code-switching: Linguistic and social perspectives for language technologies}.
\newblock In \emph{Proceedings of the 59th Annual Meeting of the Association for Computational Linguistics and the 11th International Joint Conference on Natural Language Processing (Volume 1: Long Papers)}, pages 1654--1666, Online. Association for Computational Linguistics.

\bibitem[{Felkner et~al.(2023)Felkner, Chang, Jang, and May}]{felkner2023winoqueer}
Virginia Felkner, Ho-Chun~Herbert Chang, Eugene Jang, and Jonathan May. 2023.
\newblock Winoqueer: A community-in-the-loop benchmark for anti-lgbtq+ bias in large language models.
\newblock In \emph{Proceedings of the 61st Annual Meeting of the Association for Computational Linguistics (Volume 1: Long Papers)}, pages 9126--9140.

\bibitem[{Gautam et~al.(2021)Gautam, Gupta, and Shrivastava}]{gautam-etal-2021-translate}
Devansh Gautam, Kshitij Gupta, and Manish Shrivastava. 2021.
\newblock \href {https://doi.org/10.18653/v1/2021.calcs-1.3} {Translate and classify: Improving sequence level classification for {E}nglish-{H}indi code-mixed data}.
\newblock In \emph{Proceedings of the Fifth Workshop on Computational Approaches to Linguistic Code-Switching}, pages 15--25, Online. Association for Computational Linguistics.

\bibitem[{Li et~al.(2024)Li, Fan, Atreja, and Hemphill}]{li2024}
Lingyao Li, Lizhou Fan, Shubham Atreja, and Libby Hemphill. 2024.
\newblock \href {https://doi.org/10.1145/3643829} {“hot” chatgpt: The promise of chatgpt in detecting and discriminating hateful, offensive, and toxic comments on social media}.
\newblock \emph{ACM Trans. Web}, 18(2).

\bibitem[{Lu et~al.(2023)Lu, Xu, Zhang, Min, Yang, and Lin}]{toxicn}
Junyu Lu, Bo~Xu, Xiaokun Zhang, Changrong Min, Liang Yang, and Hongfei Lin. 2023.
\newblock \href {https://doi.org/10.18653/V1/2023.ACL-LONG.898} {Facilitating fine-grained detection of chinese toxic language: Hierarchical taxonomy, resources, and benchmarks}.
\newblock In \emph{Proceedings of the 61st Annual Meeting of the Association for Computational Linguistics (Volume 1: Long Papers), {ACL} 2023, Toronto, Canada, July 9-14, 2023}, pages 16235--16250. Association for Computational Linguistics.

\bibitem[{Manikandan et~al.(2022)Manikandan, Subramanian, and Shanmugavadivel}]{manikandan2022system}
Deepalakshmi Manikandan, Malliga Subramanian, and Kogilavani Shanmugavadivel. 2022.
\newblock \href {https://ceur-ws.org/Vol-3395/T2-4.pdf} {A system for detecting abusive contents against {LGBT} community using deep learning based transformer models}.
\newblock In \emph{Working Notes of {FIRE} 2022 - Forum for Information Retrieval Evaluation, Kolkata, India, December 9-13, 2022}, volume 3395 of \emph{{CEUR} Workshop Proceedings}, pages 106--116. CEUR-WS.org.

\bibitem[{McGiff and Nikolov(2024)}]{mcgiff2024bridging}
Josh McGiff and Nikola~S Nikolov. 2024.
\newblock Bridging the gap in online hate speech detection: a comparative analysis of bert and traditional models for homophobic content identification on x/twitter.
\newblock \emph{arXiv preprint arXiv:2405.09221}.

\bibitem[{Nozza et~al.(2023)Nozza, Cignarella, Damo, Caselli, and Patti}]{HODI}
Debora Nozza, {Alessandra Teresa} Cignarella, Greta Damo, Tommaso Caselli, and Viviana Patti. 2023.
\newblock Hodi at evalita 2023: Overview of the first shared task on homotransphobia detection in italian.
\newblock In \emph{Proceedings of the Eighth Evaluation Campaign of Natural Language Processing and Speech Tools for Italian. Final Workshop (EVALITA 2023)}, CEUR Workshop Proceedings. CEUR Workshop Proceedings (CEUR-WS.org).
\newblock Publisher Copyright: {\textcopyright} 2023 Copyright for this paper by its authors. Use permitted under Creative Commons License Attribution 4.0 International (CC BY 4.0).; 8th Evaluation Campaign of Natural Language Processing and Speech Tools for Italian. Final Workshop, EVALITA 2023 ; Conference date: 07-09-2023 Through 08-09-2023.

\bibitem[{Pikuliak et~al.(2021)Pikuliak, {\v{S}}imko, and Bielikov{\'a}}]{pikuliak2021cross}
Mat{\'u}{\v{s}} Pikuliak, Mari{\'a}n {\v{S}}imko, and M{\'a}ria Bielikov{\'a}. 2021.
\newblock Cross-lingual learning for text processing: A survey.
\newblock \emph{Expert Systems with Applications}, 165:113765.

\bibitem[{Sosto and Barrón-Cedeño(2024)}]{sosto2024queerbenchquantifyingdiscriminationlanguage}
Mae Sosto and Alberto Barrón-Cedeño. 2024.
\newblock \href {https://arxiv.org/abs/2406.12399} {Queerbench: Quantifying discrimination in language models toward queer identities}.
\newblock \emph{Preprint}, arXiv:2406.12399.

\bibitem[{Tash et~al.(2023)Tash, Armenta{-}Segura, Ahani, Kolesnikova, Sidorov, and Gelbukh}]{shahiki2023lidoma}
Moein~Shahiki Tash, Jes{\'{u}}s Armenta{-}Segura, Zahra Ahani, Olga Kolesnikova, Grigori Sidorov, and Alexander~F. Gelbukh. 2023.
\newblock \href {https://ceur-ws.org/Vol-3496/homomex-paper1.pdf} {{LIDOMA} at homo-mex2023@iberlef: Hate speech detection towards the mexican spanish-speaking {LGBT+} population. the importance of preprocessing before using bert-based models}.
\newblock In \emph{Proceedings of the Iberian Languages Evaluation Forum (IberLEF 2023) co-located with the Conference of the Spanish Society for Natural Language Processing {(SEPLN} 2023), Ja{\'{e}}n, Spain, September 26, 2023}, volume 3496 of \emph{{CEUR} Workshop Proceedings}. CEUR-WS.org.

\bibitem[{Unanue et~al.(2023)Unanue, Haffari, and Piccardi}]{unanue-etal-2023-t3l}
Inigo~Jauregi Unanue, Gholamreza Haffari, and Massimo Piccardi. 2023.
\newblock \href {https://doi.org/10.1162/tacl_a_00593} {{T}3{L}: Translate-and-test transfer learning for cross-lingual text classification}.
\newblock \emph{Transactions of the Association for Computational Linguistics}, 11:1147--1161.

\bibitem[{Untermeyer(1964)}]{untermeyer1964robert}
L.~Untermeyer. 1964.
\newblock \href {https://books.google.com.au/books?id=-6paAAAAMAAJ} {\emph{Robert Frost: a Backward Look}}.
\newblock Reference Department, Library of Congress.

\end{thebibliography}
\end{document}